%% file: main.tex
\documentclass[10pt,twocolumn,letterpaper]{article}

\usepackage[pagenumbers]{cvpr} 
\usepackage{tabularx} 
\usepackage{graphicx}
\usepackage{booktabs}
\usepackage{multirow}
\usepackage{colortbl}
\usepackage{xcolor}

\definecolor{cvprblue}{rgb}{0.21,0.49,0.74}
\usepackage[pagebackref,breaklinks,colorlinks,allcolors=cvprblue]{hyperref}

\def\confName{CVPR}
\def\confYear{2025}

\title{FashionComposer: Compositional Fashion Image Generation}

\author{
    Sihui Ji$^{1,2,*}$ \quad
    Yiyang Wang$^{1}$ \quad
    Xi Chen$^{1}$ \quad
    Xiaogang Xu$^{3}$ \quad
    Hao Luo$^{2,4}$ \quad
    Hengshuang Zhao$^{1,\dagger}$\\[2pt]
    $^{1}$The University of Hong Kong \quad
    $^{2}$DAMO Academy, Alibaba Group \quad
    $^{3}$Zhejiang University \quad
    $^{4}$Hupan Lab \\
    \textit{\small \href{https://SihuiJi.github.io/FashionComposer-Page}{https://SihuiJi.github.io/FashionComposer-Page}}
}

\begin{document}
\twocolumn[{
\renewcommand\twocolumn[1][]{#1}
\maketitle
\label{fig:teaser}
\begin{center}
    \vspace{-38pt}
    \includegraphics[width=0.99\linewidth]{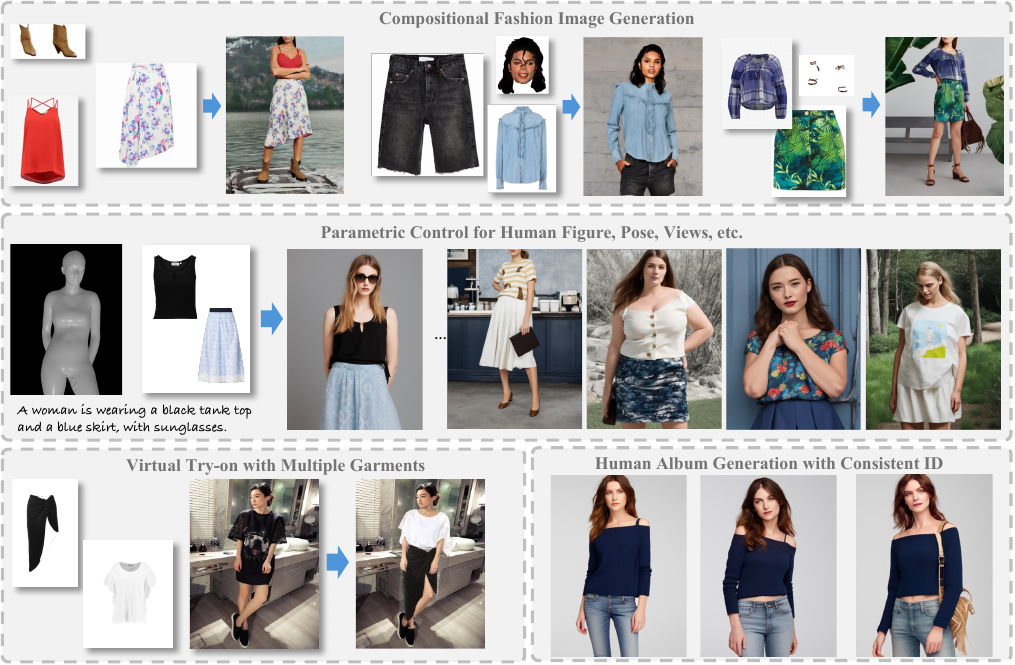}
    \vspace{-8pt}
    \captionsetup{type=figure}
    \caption{
        \textbf{Demonstration for the applications of FashionComposer.} 
        FashionComposer takes different kinds of conditions~(\textit{e.g.,} garment image, face image, parametric human model) equally as ``assets'' to composite diverse and realistic fashion images. 
        Thus supporting various fashion-related applications like controllable model image generation, virtual try-on, human album generation, \textit{etc}.  
    }
\end{center}
}]

\let\thefootnote\relax\footnotetext{*Work done when interning at DAMO Academy. $\dagger$ Corresponding author.} 

\maketitle
\input{./sec_arxiv/0_abstract} 
\input{./sec_arxiv/1_introduction}
\input{./sec_arxiv/2_related_work}
\input{./sec_arxiv/3_methods}
\input{./sec_arxiv/4_experiments}
\input{./sec_arxiv/5_conclusion}
{
    \small
    \bibliographystyle{ieeenat_fullname}
    \bibliography{main}
}


\end{document}

%% file: sec_arxiv/0_abstract.tex
\begin{abstract}
We present \textbf{FashionComposer} for compositional fashion image generation.
Unlike previous methods, FashionComposer is highly flexible.
It takes multi-modal input~(\textit{i.e.,} text prompt, parametric human model, garment image, and face image) and
supports personalizing the appearance, pose, and figure of the human and assigning multiple garments in one pass. 
To achieve this, we first develop a universal framework capable of handling diverse input modalities. 
We construct scaled training data to enhance the model's robust compositional capabilities. 
To accommodate multiple reference images~(garments and faces) seamlessly, we organize these references in a single image as an ``asset library'' and employ a reference UNet to extract appearance features.
To inject the appearance features into the correct pixels in the generated result, we propose subject-binding attention.
It binds the appearance features from different ``assets'' with the corresponding text features. 
In this way, the model could understand each asset according to their semantics, supporting arbitrary numbers and types of reference images.
As a comprehensive solution, FashionComposer also supports many other applications like human album generation, diverse virtual try-on tasks, etc.
\vspace{-5pt}
\end{abstract}

%% file: sec_arxiv/1_introduction.tex
\vspace{-8pt}
\section{Introduction}
\label{sec:intro}
In the e-commercial age, the fashion industry is inundated with large amounts of in-shop garment images. To better attract customers, they hire models to showcase how the clothes look on individuals. 
Virtual try-on technology, designed to generate an image of the specific human wearing provided garments, has become increasingly popular.
Existing virtual try-on methods~\cite{xie2023gp, choi2024improving,kim2024stableviton,wang2024mv} usually allow for trying on only a single garment and have significant limitations in terms of flexibility.
For example, the poses of the synthesized images are fixed according to the human image, which imposes strict constraints on the diversity of body shapes and postures in try-on images.
Furthermore, these methods are usually conditional on only one garment, thus are unable to try on outfits.

Facing the aforementioned challenges, we introduce FashionComposer, a flexible fashion image generator. 
Unlike existing strategies, the core feature of FashionComposer is \textit{compositionality}, manifested in two aspects. 
First, it involves \textit{multi-modal inputs}, including a language description for the target fashion image, a parametric human model that controls the human figure and posture, and garment images (face as optional) for reference. 
Second, it entails the composition of \textit{multiple visual assets}.
It allows users to drag their desired garment components and human faces into an ``asset library'' to customize the generation.

To achieve such compositionality, we design a diffusion-based framework with multi-modal inputs. 
We employ the Skinned Multi-Person Linear model~(SMPL)~\cite{loper2023smpl} to control the human figure and posture and leverage a series of off-the-shelf models~\cite{bai2023qwenvl, guler2018densepose, cheng2021mask2former} to prepare multi-modal training data~(\textit{e.g.,} the masks and language descriptions for different parts). 
To handle multiple visual assets~(\textit{e,g.,} different garments, face, shoes) in one pass without increasing the computation burden, we propose to arrange all the referring elements in one image as an ``asset library'' and leverage one reference UNet~\cite{hu2023animate} to extract the features.
Considering the asset image might contain different numbers of garment images with different positions and sizes, it is hard for the model to understand the mapping relation between the asset image and the generation result. 
To solve this issue,  we propose subject-binding attention, which binds the appearance features of the garments/faces with the corresponding text embeddings.
Therefore, the features of each asset could be mapped to the correct pixels according to their semantics.

Besides pursuing composability, we further develop correspondence-aware attention and latent code alignment to support FashionComposer in generating an album of images with a consistent identity. 
As an all-round fashion image generator, FashionComposer also supports traditional virtual try-on and multi-garment virtual try-on tasks.
In general, FashionComposer demonstrates an extraordinary capacity for multi-modal fashion generation. Our contributions could be summarized in three folds.
\begin{itemize}
    \item We propose FashionComposer, which takes multi-modal input conditions and supports multiple fashion-related tasks in a unified framework.  
    \item We design subject-binding attention, which enables the composition of multiple visual assets in an effective and extensible manner. 
    \item We introduce techniques like correspondence-aware attention and latent code alignment to support multiple practical applications like album image generation, multi-garment try-on, \textit{etc}.
\end{itemize}

%% file: sec_arxiv/2_related_work.tex
\vspace{-5pt}
\section{Related Work}
\label{sec:Related Work}
\vspace{-5pt}

\textbf{Standard virtual try-on.}
The goal of the standard virtual try-on task is to generate images where a garment from an in-shop image is seamlessly integrated onto a reference person while preserving meticulous details.
Warping-based methods~\cite{han2017viton,xie2023gp,wang2018toward} depend on garment deformation followed by fitting on the individual to generate the composite images, suffering from alignment issues. Diffusion-based methods can directly dress a target person in
a specified garment without warping based on the strong ability of generative models. For maintenance
of identity, LaDIVTON~\cite{morelli2023ladi} uses CLIP~\cite{radford2021learning} as image encoder for feature extraction while TryOnDiffusion~\cite{zhu2023tryondiffusion} designs a two-UNets architecture, and OOTDiffusion~\cite{xu2024ootdiffusion} leverages reference UNet for maintaining details of clothes.
However, a notable limitation of these methods is their restricted capacity for customization. Specifically, they are constrained to transferring only a single garment onto the reference person, hindering flexibility in the try-on process.

\noindent \textbf{Multi-object customization.}
Discerning different objects while preserving their identities is a challenging task.
Textual Inversion~\cite{gal2023image} and DreamBooth~\cite{ruiz2023dreambooth} propose to tune the text representation space to add new concepts for various objects in the text-to-image diffusion models.
Cones~\cite{liu2023cones} finds the neurons in the neural network of each corresponding referred object.
Emu2~\cite{sun2023generative} is a multi-modal autoregressive model that integrates the input text prompt and different reference objects in a multi-modal sequence and generates images accordingly.
Collage Diffusion~\cite{sarukkai2024collage} generates multiple objects by separating each object into different layers, which can transform a user-provided object layout into a high-resolution image. 
FastComposer~\cite{xiao2023fastcomposer} enables multi-person composition by binding the image embeddings of different people to the corresponding word embeddings, which can synthesize personalized, multi-person images without additional tuning.
Nevertheless, these techniques either demand costly fine-tuning~\cite{sarukkai2024collage, gal2023image, ruiz2023dreambooth, liu2023cones}, or struggle to keep detailed fidelity~\cite{sarukkai2024collage, ruiz2023dreambooth, xiao2023fastcomposer, gal2023image, sun2023generative}. 
In contrast, FashionComposer can generate multi-garment fashion images without tuning and keep detailed fidelity for each garment with excellent superiority.

%% file: sec_arxiv/3_methods.tex
\begin{figure*}[t]
\centering 
\includegraphics[width=0.99\linewidth]{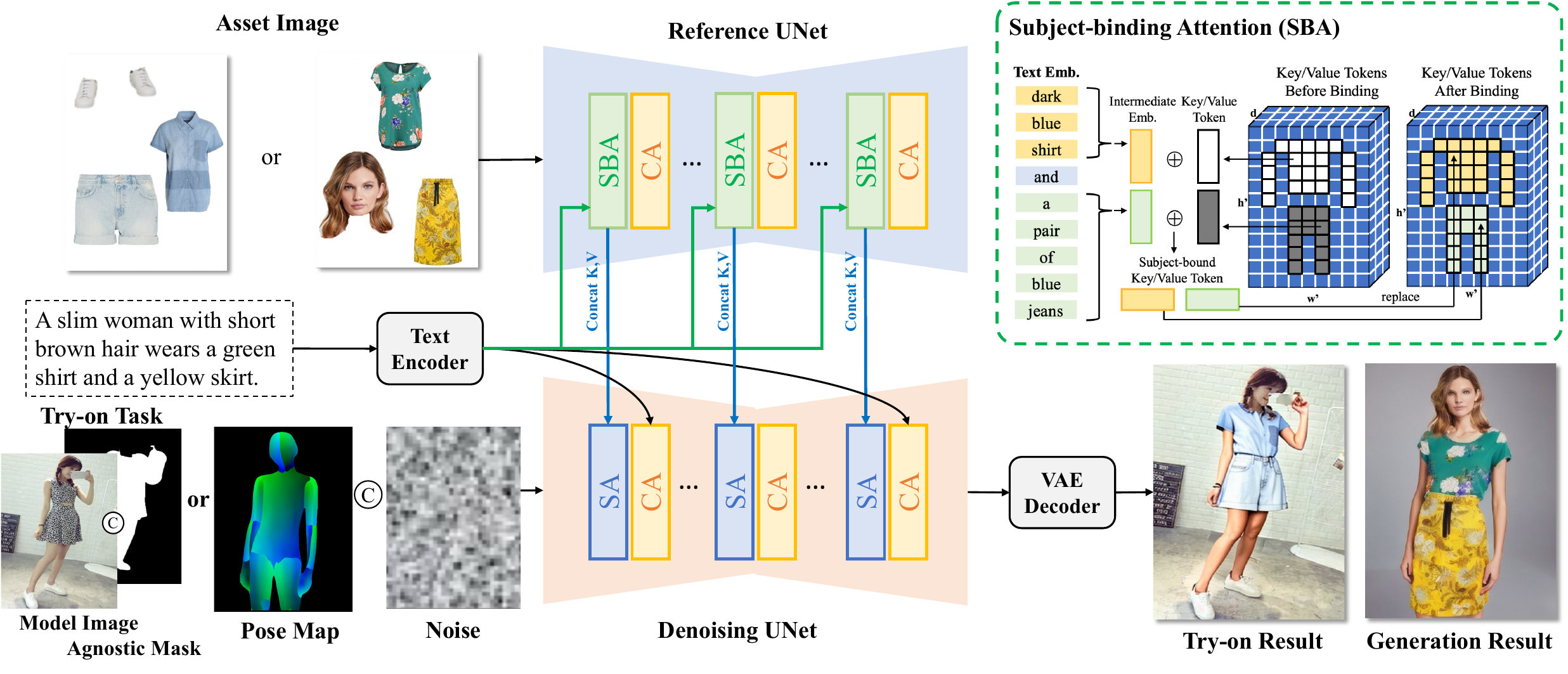} 
\vspace{-10pt}
\caption{%
   \textbf{Overall pipeline of FashionComposer.}
    FashionComposer takes garments composition and optional face, text prompt, and a densepose map projected from SMPL as inputs. The text prompt is encoded and fused with UNets through cross-attention and subject-binding attention, while the garment features are extracted and injected for denoising through Feature Injection Attention. 
}
\label{fig:pipeline}
\vspace{-10pt}
\end{figure*}

\section{Method}
\label{sec:Methods}
\subsection{Preliminaries}
\textbf{Stable Diffusion.}
We use Stable Diffusion (SD) v1.5~\cite{rombach2022high} as our backbone model. The UNet-based SD model comprises a variational autoencoder(VAE), a denoising UNet, and a text encoder. The VAE encoder $\mathcal{E}$ encodes the image $x$ into a compressed latent representation $z$. The latent $z$ is then perturbed by a Gaussian noise $\epsilon$ in the forward diffusion process, which is predicted by the UNet parameterized by $\theta$ for the backward diffusion process. The text prompt $\mathcal{P}$ is encoded by the text encoder $\mathcal{T}$ into text embeddings $\mathcal{T(P)}$, which is then injected into the UNet through cross-attention. During training, the network is optimized to minimize the loss function formulated as:
\begin{align}
    \mathcal{L}=\mathbb{E}_{t\sim \mathbf{T}, \epsilon \sim \mathcal{N}(0,1), z\sim \mathcal{E}(x)}[||\epsilon - \epsilon_{\theta}(z_t,t,\mathcal{T(P)})||^2_2],
\end{align}
where the timestep $t$ is sampled by a timestep scheduler $\mathbf{T}$, and $z_t$ is the latent representation at the timestep $t$. During inference, the initial noise $z_T$ is sampled from $\mathcal{N}(0,1)$ and iteratively denoised by the denoising UNet that predicts the noise at each timestep conditioned on the text prompt $\mathcal{P}$. The final latent code $z_0$ is the input of the VAE decoder $\mathcal{D}$ to generate the final image $x_0 = \mathcal{D}(z_0)$.

\subsection{Framework Overview}

The overall pipeline of FashionComposer is demonstrated in \cref{fig:pipeline}. 
The densepose map projected from SMPL  is initially concatenated with the noise, followed by the denoising UNet performing the denoising process. 
The reference UNet follows the same framework as the denoising UNet, with the distinction that the self-attention modules are substituted with subject-binding attention.

FashionComposer leverages different combinations of input to support different tasks, naturally formulating fashion image generation and conventional virtual try-on in the same framework.
The only change we make to fit the
original inpainting-based virtual try-on setting is to concatenate a 4-channel cloth-agnostic person image embedding encoded by VAE and a 1-channel binary cloth-agnostic mask along with noise rather than using a densepose map. The convolution-in channel of UNet is converted to 9 accordingly.
FashionComposer achieves the leading performance for compositional fashion image generation and our finetuned virtual try-on model achieves superior qualitative and quantitative results on standard try-on benchmarks while supporting multiple garment try-on in one pass.

\subsection{Composition for Multi-modal Conditions}
\noindent \textbf{Multi-modal input conditions.}
Unlike existing virtual try-on methods relying on only image inputs, FashionComposer seamlessly integrates various multi-modal conditions (\textit{i.e.}, text prompt, SMPL parameters, reference garments and optional face) in the fashion image generation. We condition the generator on different multi-modal conditions simultaneously using a unified framework. 
For the \textit{visual assets composition}, we employ the reference UNet to extract its multi-level features, which are then involved in the denoising process by attention sharing.
As for the \textit{text prompt}, we encode it into text embeddings using CLIP's text encoder. We opt to condition them on the cross-attention modules of both UNets for appearance guidance.
Additionally, we opt for a lighter approach in handling the \textit{SMPL} , as we aim to avoid introducing overly intricate conditions into the generation process, which would compromise the generation quality. Therefore, we choose to simply concatenate the 2-channel densepose map projected from SMPL with the noisy latent along the feature channel and directly input them into the denoising UNet. 

Adopting this approach, we effectively condition FashionComposer on diverse multi-modal inputs. FashionComposer demonstrates extraordinary abilities to follow different conditions and synthesize fashion images accordingly.

\noindent \textbf{Multi-modal data construction.}
\label{sec: dataset_construction}
The ideal training samples consist of all in-shop garments and the dressed human images, along with text descriptions, and an index indicating the correlation between phrases in the text and the corresponding garments. However, the raw datasets we utilize~\cite{han2017viton, morelli2022dresscode, liuLQWTcvpr16DeepFashion} contain only a single corresponding in-shop garment image for each target image.

To address this problem, we use the Mask2Former-Parsing~\cite{yang2023humanparsing, cheng2021mask2former} to detect the human parsing maps of the person images. Then for our chosen garments without a corresponding in-shop garment image, we mask them out as the selected components of the garment's composition. 
Subsequently, we construct the composition of in-shop garment, masked-out garment and face by randomly placing them without overlapping on a white background. We leverage Qwen-VL-Chat~\cite{bai2023qwenvl} to caption the target images. For each text prompt, we identify the phrases describing different components and classify them into corresponding categories by querying Qwen-14B-Chat~\cite{bai2023qwen} about the phrases related to each component in the sentence. The final joint multi-modal dataset includes 165k samples.

\subsection{Composition of Multiple Visual Assets}
Preserving reference garments' global features and intricate details is crucial for ensuring the fidelity of the synthesized images. 
Reference UNet~\cite{hu2023animate,chen2024zero, chen2024wear,zhang2024flashface} is proven effective for preserving the fine details of the reference image. 
In this work, we follow this structure to extract the reference appearance features and design subject-binding attention to enable multiple references in one pass.

\noindent \textbf{Basics for reference UNet.}  Specifically, the reference UNet follows the same framework as the denoising UNet initialized with the parameters from SD V1.5. The asset composition undergoes encoding by the reference UNet, which converts it into multi-level features across the different blocks of the UNet. 
Specifically, we extract the key and value tokens from the self-attention modules of the reference UNet, which can be formulated as $k_{ref},v_{ref} \in \mathbb{R}^{(h\times w)\times d}$. Then we concatenate these tokens with their corresponding key and value tokens in the matching block of the denoising UNet, which can be formulated as $[k_{den},k_{ref}],[v_{den},v_{ref}] \in \mathbb{R}^{(2\times h\times w)\times d}$ (the operator $[x_1,x_2,...]$ stands for concatenation). Consequently, the self-attention operations in the denoising UNet are replaced with a newly defined attention
formulated as:
\begin{align}
    \text{softmax}(\frac{q_{den} \cdot [k_{den},k_{ref}]^T}{\sqrt{d}})\cdot [v_{den},v_{ref}].
\end{align}

Note that different from \cite{hu2023animate}, we keep textual input
for the cross-attention module.
While reference UNet utilizes the existing feature modeling capacities of the pre-trained SD for accurately maintaining details, the model still retains the ability to align with the text prompt which facilitates the generation process under
multiple guidance.

\noindent \textbf{Subject-binding attention.}
A limitation of standard reference UNet is that it cannot deal with multiple reference images. 
A direct approach is equipping each reference image with a specific reference UNet. However, this approach is computationally expensive, especially for cases with large numbers of references.
Facing this challenge, we propose subject-binding attention. Specifically, we support users to put all their reference elements~(with arbitrary location, size, or number) in one asset image. Then, we let one reference UNet to extract the features for this asset image. Afterward, to make the model understand the ``all-in-one'' features, we bind each pixel with their corresponding text descriptions according to the following steps:   

First, we find the corresponding tokens of each visual asset in different feature maps.
Suppose we select the UNet block $l$ here; then, the size of the feature map in this block is determined, denoted by $h'\times w'$. Then, given an asset's composition $G=\{a_1,a_2,...a_n\}$, for each asset component $a_i$ in the asset's composition, we directly downsample it to derive its corresponding area in the feature map. 
Since the key/value tokens in the self-attention module can be seen as feature maps flattened into sequences, denoted by $K\in \mathbb{R}^{(h'\times w')\times d}, V\in \mathbb{R}^{(h'\times w')\times d}$, then the above downsample operation determines $n_i$ corresponding key/value points $K_{i} = \{k_1,k_2...k_{n_i}\}, V_{i} = \{v_1,v_2...v_{n_i}\}$ for asset component $a_i$ in the self-attention module. 

Second, we bind the selected points (tokens) with their corresponding text representations.
Take the key points as an example. Each key point $k_j\in \mathbb{R}^{d}$ is an element of the key token $K$. Now, we proceed to bind all key points in $K_i$ with the text embeddings corresponding to asset $a_i$.
The text prompt is denoted by $\mathcal{P}=\{w_1,w_2,...,w_s\}$, then for asset $a_i$ it might have a corresponding phrase $\{w_{i_1},...w_{i_p}\}$ like ``a red shirt'', whose text embedding is denoted as $\mathcal{P}_{i} \in \mathbb{R}^{l \times c}$. We input this text embedding into an MLP, with each UNet block having one, and then add the output with each $k_j$ in $K_i$, formulated as:
\begin{align}
  k'_j = \text{MLP}_{l}(\mathcal{P}_i) + k_j.
\end{align}

Subsequently, we replace each original key point $k_j\in K_i$ in the key token with the subject-bind embedding $k'_j$. This procedure is repeated for each asset in the asset's composition, and same 
for the value tokens. \cref{fig:pipeline} illustrates the concrete procedure of our approach which provides the model with awareness of the semantic meaning of conditioned asset to better maintain details without confusion.

\subsection{Consistent Human Image Generation}
Another feature of FashionComposer is generating a human album with a consistent ID. Specifically, we propose correspondence-aware attention and latent code alignment.

\noindent\textbf{Correspondence-aware attention.} Inspired by cross-frame attention used in text-to-video area~\cite{khachatryan2023text2video}, we first try to directly replace all key/value tokens of self-attention modules of the $2^{nd}$ to $N^{th}$ images with the key/value tokens of the $1^{st}$ image. Although the synthesized human appearances demonstrate excellent consistency, the fidelity and quality of the garments are compromised. We attribute this to the excessive amount of information from the first image, which negatively impacts the subsequent images. Therefore, we propose correspondence-aware attention to leverage the information from the first image and the current image. 
Specifically, we extract the UV map coordinates for all SMPL parameters and map them on the densepose maps, where identical UV coordinates $(u,v)$ denote the same human area across different SMPL settings. Then, we substitute the key/value tokens of the $2^{nd}$ to the $N^{th}$ image with those of the $1^{st}$ image only if they share the same $(u,v)$ coordinates. While such a design effectively generates an album of high-fidelity images, it exhibits poorer consistency in human appearances compared to cross-frame attention.

\noindent\textbf{Latent code alignment.}
To enhance the consistency in human appearances, we propose latent code alignment to capitalize on the high consistency provided by cross-frame attention and the high fidelity offered by correspondence-aware attention. Specifically, we first generate an album of images using cross-frame attention and preserve their denoised latent codes $\{z_0^1,z_0^2,...z_0^N\}$ in the final denoising timestep. Next, we utilize the densepose map to extract face masks representing facial areas across different images. We then apply these masks to remove the facial regions from these denoised latent codes and stitch them into the corresponding facial areas of the latent codes synthesized by correspondence-aware attention. This design does not require additional fine-tuning and synthesizes an album of a given human with both high consistency and fidelity.

%% file: sec_arxiv/4_experiments.tex
\vspace{-5pt}
\section{Experiments}
\subsection{Implementation Details}

\noindent\textbf{Hyperparameters.}
During training, we rescale the image resolution to $512\times384$. We choose the AdamW optimizer~\cite{loshchilov2017decoupled} with an initial learning rate of $1e^{-4}$. 
The trainable modules are the self-attention modules, cross-attention modules of the denoising UNet and the reference UNet, the MLP used in text augmentation, and the convolution-in layer of the denoising UNet.

\noindent\textbf{Evaluation metrics.}
We evaluate the similarity between the synthesized and the reference garments by calculating the CLIP-Score (CLIP-I) and DINO-Score following DreamBooth~\cite{ruiz2023dreambooth}. Furthermore, we evaluate the prompt consistency by calculating the CLIP text-image similarity (CLIP-T) following Textual Inversion~\cite{gal2023image}.
Additionally, we organize user studies with a group of 23 annotators to compare the generation results in quality and fidelity.

\begin{figure}[t]
    \centering
    \includegraphics[width=1.0\linewidth]{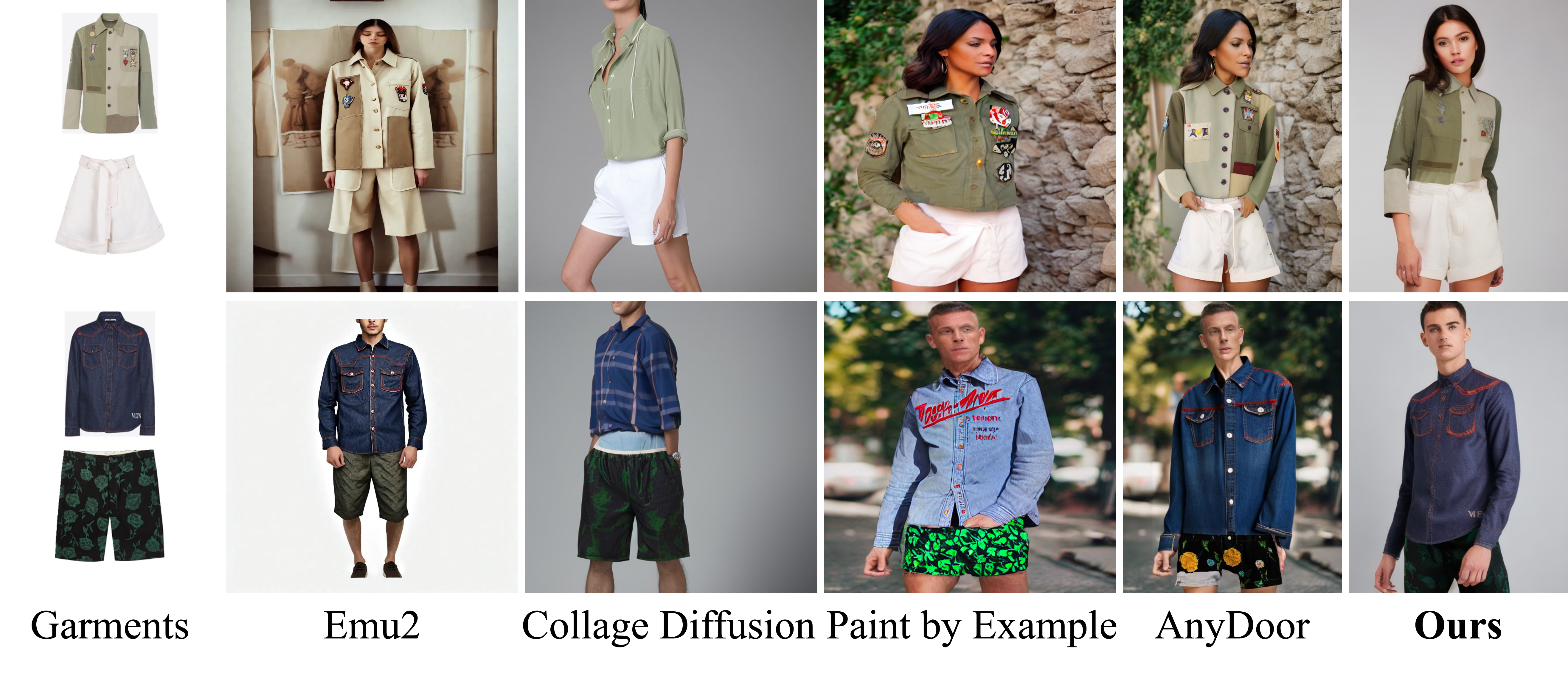}
    \vspace{-20pt}
    \caption{\textbf{Qualitative comparison with multi-reference customization methods}, including Emu2~\cite{sun2023generative}, Collage Diffusion~\cite{sarukkai2024collage}, Paint by Example~\cite{yang2023paint} and AnyDoor~\cite{chen2024AnyDoor}. 
    }
    \label{fig: multi_comp}
    \vspace{-10pt}
\end{figure}

\begin{figure*}[t]
\centering 
\includegraphics[width=1.0\linewidth]{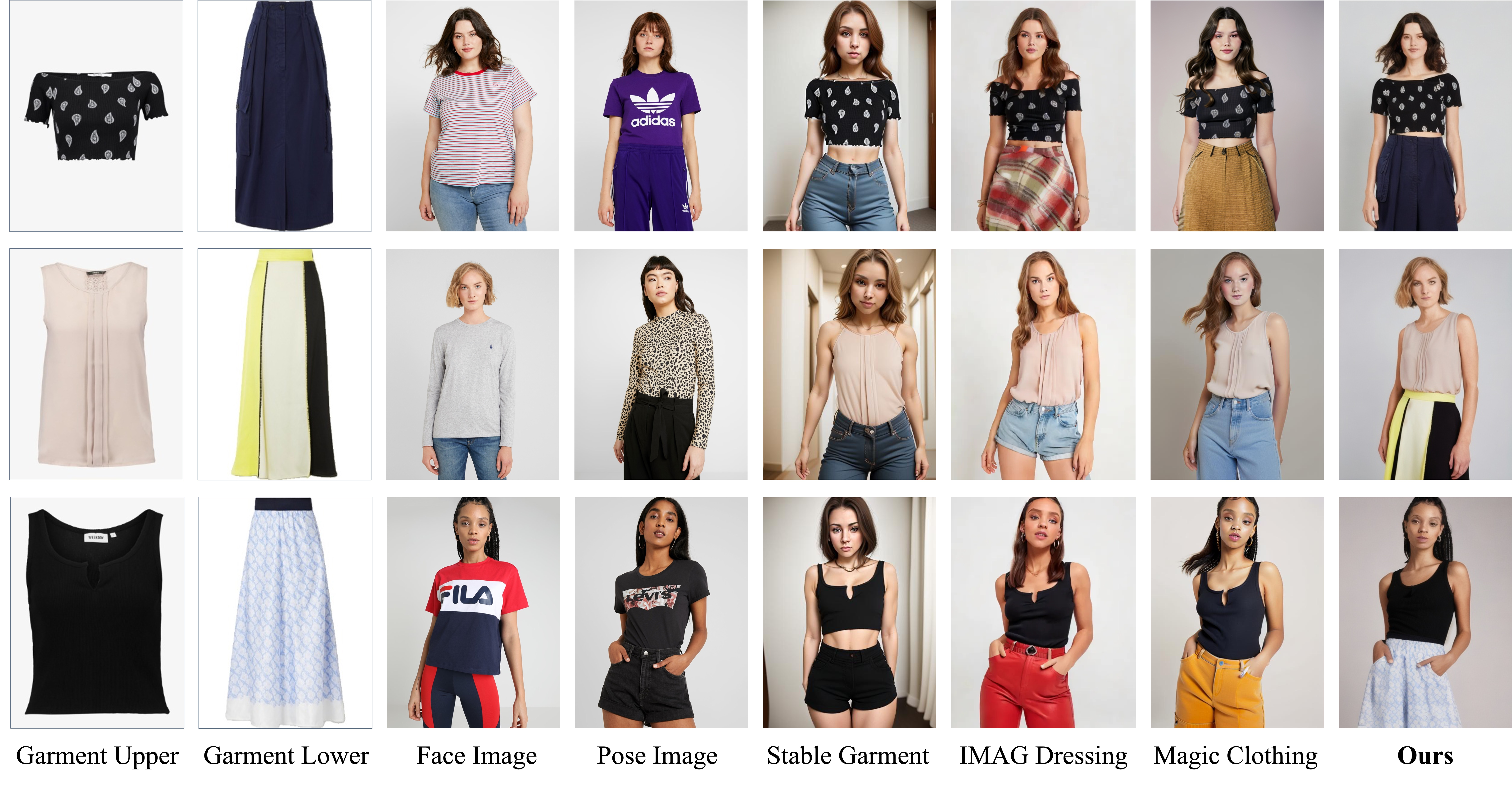} 
\vspace{-20pt}
\caption{%
    \textbf{Qualitative comparison with garment-centric fashion image synthesis methods}, including
    StableGarment~\cite{wang2024StableGarment},
    IMAGDressing-v1~\cite{shen2024IMAGDressing-v1},
    and Magic Clothing~\cite{chen2024MagicClothing},
    where ours better preserves the identity of the target objects.
    Note that all approaches do not finetune the model on the test samples.}
\label{fig:multi_comp2}
\vspace{-15pt}
\end{figure*}

\subsection{Comparisons for Compositional Generation}
We assess the fidelity-maintenance capability of our method in generating multi-guidance fashion images. Considering that only a few existing works match our task setting, we implement the comparison with two groups of methods.

\noindent  \textbf{General customization methods.} To evaluate our generation ability conditioned on multi-garment, we compare with the leading general customization methods which receive multiple subjects for reference, comprising Emu2~\cite{sun2023generative} and Collage Diffusion~\cite{sarukkai2024collage}. We also choose AnyDoor~\cite{chen2024AnyDoor} and Paint by Example~\cite{yang2023paint} as a two-stage reference-based generation method, for it requires a pre-generated background image to place the reference objects. Specifically, we first leverage a text-to-image model SD V1.5 with densepose-ControlNet to customize a background human, and then we place the reference garments in multiple passes.

In qualitative comparisons, we adapt the inputs correspondingly and make our best effort to adjust each method for better results as presented in \cref{fig: multi_comp}. 
Collage Diffusion~\cite{sarukkai2024collage} tends to blend the identity of different garments and encounters difficulties in maintaining detailed fidelity. While Emu2~\cite{sun2023generative} demonstrates improved ability in discerning different garments, it still faces challenges in preserving fidelity. AnyDoor~\cite{chen2024AnyDoor} and Paint by Example~\cite{yang2023paint} generate images with better fidelity, but they both rely on the text-to-image model to create a target human and necessitate inputting different garments in multiple forwards. 
In contrast, FashionComposer demonstrates promising performance for one-pass multi-reference customization with the best detail fidelity and an excellent understanding of garment subjects.

In quantitive comparisons, we first prepare 100 multi-garment reference images and corresponding prompts as input conditions. For multi-reference customization methods Emu2 and Collage Diffusion, we customize fashion images through one pass with possible additional input like bounding boxes for Emu2. For single reference-based methods including AnyDoor and Paint by Example, we need to generate densepose-conditioned person images and their cloth-agnostic image by utilizing Controlnet and human parsing model. Then we can implement the two-stage inpainting pipeline for image customization under multi-garment guidance. CLIP-I, CLIP-T, and DINO are used as the evaluation metrics for comparisons of image similarity between reference images and customized images as well as text-image similarity between text prompts and fashion images.
\cref{tab:quan_comp} shows that our method outperforms all compared methods in both text alignment and image similarity.

\begin{table}[t]
  \setlength{\belowcaptionskip}{0cm}
  \renewcommand{\arraystretch}{0.9} 
  \caption{\textbf{Comparison with multi-object reference generation methods.} The first three rows represent one pass multi-reference customization methods and the last two rows represent two stage inpainting pipeline based on pre-generated base images.}
  \label{tab:quan_comp}
  \vspace{-5pt}
  \centering
  \footnotesize
  \begin{tabularx}{0.45\textwidth}{@{}lXXX@{}}
    \toprule
    Method & CLIP-I$\uparrow$ & DINO$\uparrow$ & CLIP-T$\uparrow$ \\
    \midrule
    \textbf{Ours}  & \textbf{77.60}  & \textbf{40.11}  & \textbf{27.71}  \\
    Emu2        & 69.70  & 35.96  & 20.54  \\
    Collage Diffusion  & 67.80  & 34.16  & 22.14  \\
    \midrule
    AnyDoor+ControlNet & 72.40 & 37.94 & 27.00 \\
    Paint-by-example+ControlNet & 64.50 & 34.60 & 23.77 \\
    \bottomrule
    \end{tabularx}
    \vspace{-10pt}
\end{table}

\noindent \textbf{Fashion image synthesis.} To evaluate our garment-driven customization ability with the condition of the face and pose image, we compare with recent garment-centric methods which receive garment image and text description as key conditions and images of face and pose as possible additional references. 

As demonstrated in \cref{fig:multi_comp2},
all the generation methods keep high fidelity of fine-grained garment details while none of them can achieve highly flexible compositionality as our work does. StableGarment~\cite{wang2024StableGarment} could only be guided by textual description and upper cloth. IMAGDressing-v1~\cite{shen2024IMAGDressing-v1} and Magic Clothing~\cite{chen2024MagicClothing} can take all conditions except for lower cloth as guidance to generate a target person with a specified pose, face, and upper cloth. However, they run into trouble when keeping face identities and taking lower garments as references. 

\begin{figure}[t]
    \centering
    \includegraphics[width=1.0\linewidth]{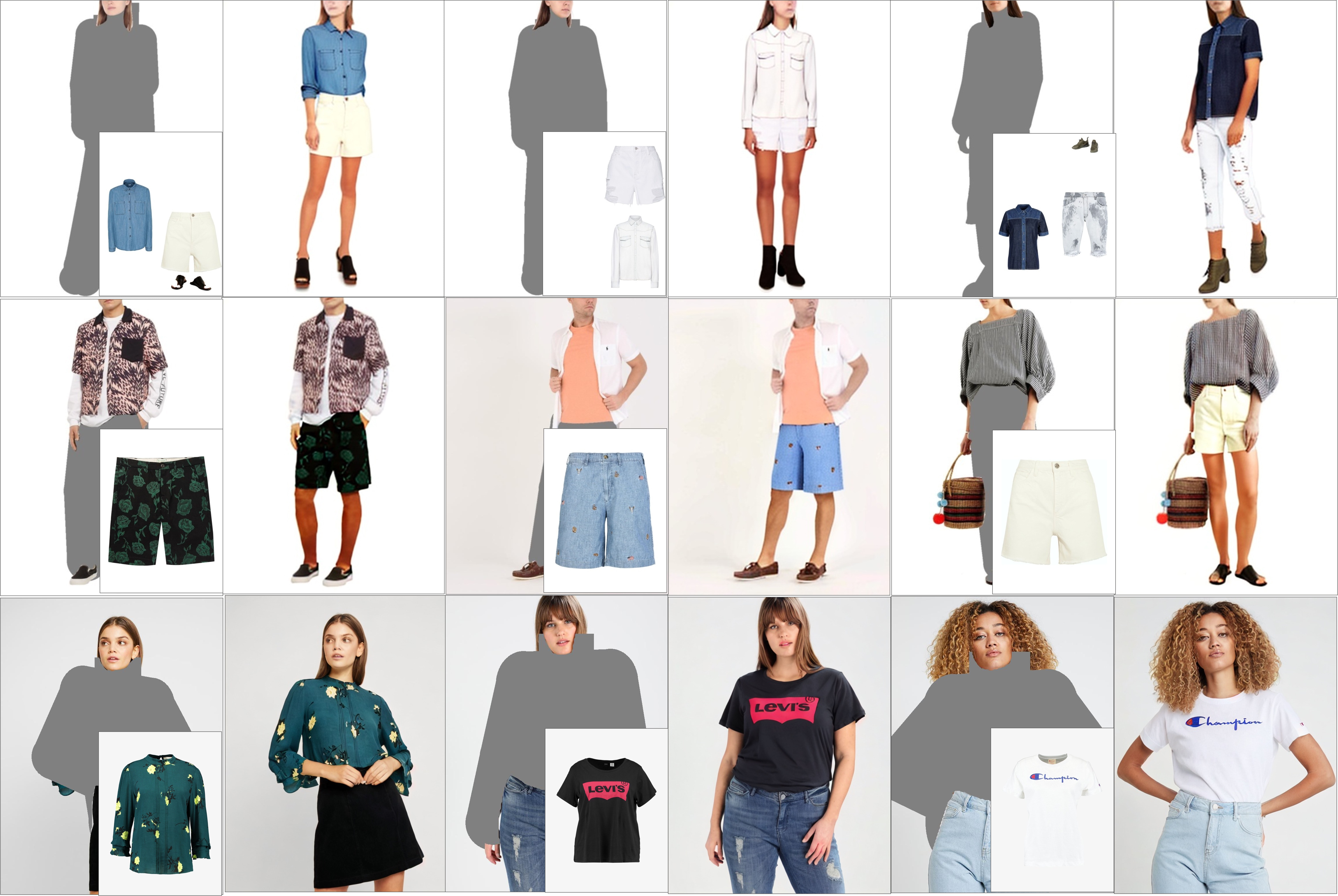}
    \vspace{-20pt}
    \caption{\textbf{Diverse virtual try-on results} of FashionComposer for upper, lower, and outfit try-on tasks.
     }
    \label{fig: tryon}
    \vspace{-5pt}
\end{figure}

\begin{figure}[t]
    \centering
    \footnotesize
    \includegraphics[width=1.0\linewidth]{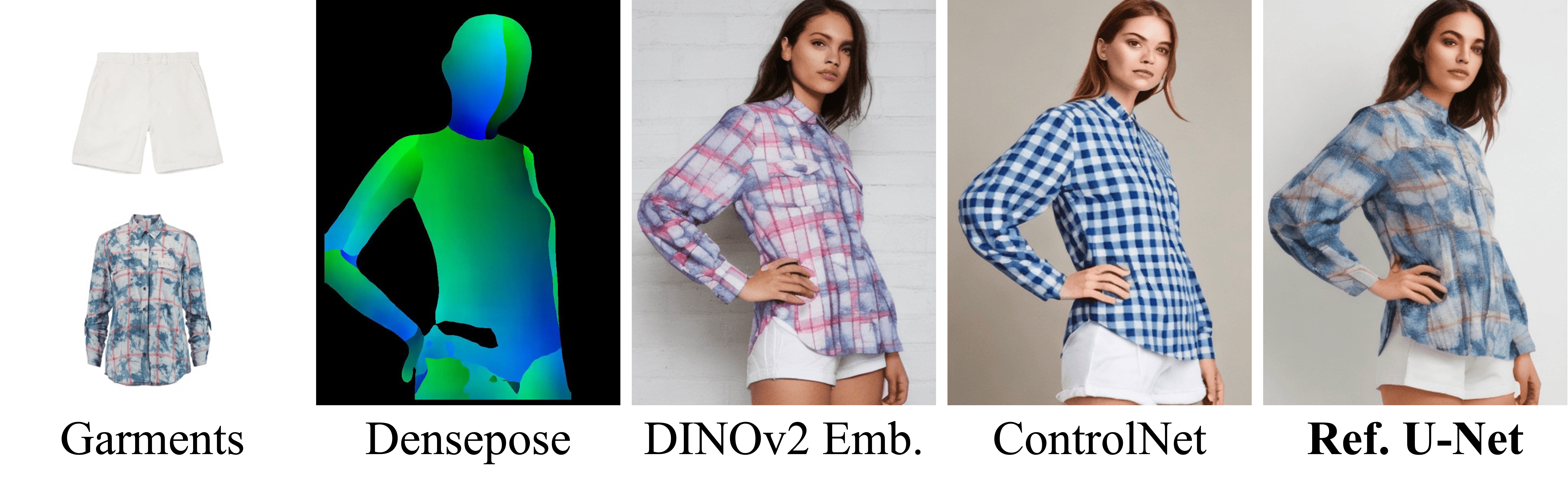}
    \vspace{-20pt}
    \caption{\textbf{Qualitative comparison for the reference encoder.} Reference UNet better preserves the fine details of the garments.} 
    \label{fig: ablation_reference}    
    \vspace{-15pt}
\end{figure}

\subsection{Comparisons of Virtual Try-on}
\noindent \textbf{Standard try-on.} A quantitative comparison is conducted on VITON-HD~\cite{han2017viton} dataset with several state-of-the-art open-source virtual try-on methods.
The comparison is performed under both paired and unpaired settings to measure the similarity between the synthesized results and ground truth and the generalization
performance of the models. The results are presented in
\cref{tab:tryon} and our model outperforms all others across all metrics except LPIPS. GP-VTON and DCI-VTON as warping-based methods, have advantages in SSIM and LPIPS but perform weaker in KID and FID. This result suggests that warping-based methods may focus more on ensuring structural and perceptual similarity but lack realism and detailed naturalness than our diffusion-based model, which convincingly verifies that our method can successfully preserve the slight details and achieve a more realistic try-on effect for these garments. 

\noindent \textbf{Multi-garment try-on.} \cref{fig: tryon} illustrates our results on various virtual try-on tasks, including upper garment try-on conducted on VITON-HD~\cite{han2017viton} dataset, and lower garment try-on as well as outfit try-on on DressCode~\cite{morelli2022dresscode} dataset. For uppers and lowers, our approach can generate flawless images preserving the consistent identity of the garment textures and devoid of artifacts.
Regarding outfit try-on setting, our method can accurately recognize the type of elements in asset image even when shoes are included, and generate reasonable results with precise fitting between garment and body parts.

\subsection{Ablation Study}

\label{sec: ablation}
\textbf{Reference UNet.}
We ablate different techniques to keep the detail fidelity of the reference garments. For DINOv2 embeddings of the garments, we condition them on the denoising UNet through cross-attention and omit the reference UNet and the text prompt. For ControlNet, we input the garment's composition into the ControlNet branch in place of the reference UNet. 

\cref{fig: ablation_reference} presents the qualitative results of different techniques. All methods succeed in aligning the posture of the synthesized individual with the reference densepose map. In terms of detail fidelity, ControlNet can only grasp the main semantic information of the garments and it fails to maintain the fidelity. Conditioning the denoising UNet on DINOv2 embeddings improves the fidelity, but it still struggles with garments with more intricate textures and details. The Reference UNet demonstrates a significant advancement in maintaining detail fidelity. It faithfully preserves all the patterns, textures, and tiny details of each reference garment.
\cref{tab:ablationRef} presents the quantitative ablation study on the effectivity of the reference UNet. Reference UNet surpasses other methods in image similarity (CLIP-I and DINO) and text-image similarity (CLIP-T).

\begin{table}[t]
  \setlength{\belowcaptionskip}{0cm}
  \renewcommand{\arraystretch}{0.9} 
  \caption{\textbf{Quantitative comparison for the standard virtual try-on task} on the VITON-HD test dataset.}
  \label{tab:tryon}
  \vspace{-5pt}
  \centering
  \resizebox{0.48\textwidth}{!}{ 
  \begin{tabular}{lccccc|cc}
    \toprule
    \multirow{2}{*}{Methods} & \multicolumn{4}{c}{VITON-HD Paired} & \multicolumn{2}{c}{Unpaired} \\
    \cmidrule(lr){2-5} \cmidrule(lr){6-7}
     & SSIM $\uparrow$ & FID $\downarrow$ & KID $\downarrow$ & LPIPS $\downarrow$ & FID $\downarrow$ & KID $\downarrow$ \\
    \midrule
    DCI-VTON~\cite{gou2023taming}  & 0.8620 & 9.408 & 4.547 & 0.0606 & 12.531 & 5.251 \\
    StableVITON~\cite{kim2024stableviton} & 0.8543 & 6.439 & 0.942 & 0.0905 & 11.054 & 3.914 \\
    StableGarment~\cite{wang2024StableGarment}  & 0.8029 & 15.567 & 8.519 & 0.1042 & 17.115 & 8.851 \\
    MV-VTON~\cite{wang2024mv} & 0.8083 & 15.442 & 7.501 & 0.1171 & 17.900 & 8.861 \\
    GP-VTON~\cite{xie2023gp}  & 0.8701 & 8.726 & 3.944 & \textbf{0.0585} & 11.844 & 4.310 \\
    LaDI-VTON ~\cite{morelli2023ladi}& 0.8603 & 11.386 & 7.248 & 0.0733 & 14.648 & 8.754 \\
    OOTDiffusion~\cite{xu2024ootdiffusion} & 0.8187 & 9.305 & 4.086 & 0.0876 & 12.408 & 4.689 \\
    \textbf{Ours} & \textbf{0.8771} & \textbf{5.842} & \textbf{0.906} & 0.0727 & \textbf{9.205} & \textbf{1.3606} \\
    \bottomrule
    \end{tabular}
    }
    \vspace{-8pt}
\end{table}

\begin{table}
  \setlength{\belowcaptionskip}{0cm}
  \renewcommand{\arraystretch}{0.9} 
  \caption{\textbf{Quantitative study for the reference UNet.} We compare with other options for the appearance encoders like DINOv2 and ControlNet. Reference UNet shows the best performance.}
  \label{tab:ablationRef}
  \vspace{-5pt}
  \footnotesize
  \centering
  \begin{tabularx}{0.49\textwidth}{@{}lXXX@{}}
    \toprule
    Method & CLIP-I$\uparrow$ & DINO$\uparrow$ & CLIP-T$\uparrow$ \\
    \midrule
    DINOv2 Embeddings & 76.80 & 38.22 & 26.17\\
    ControlNet & 75.94 & 33.47 & 27.10\\
    Reference UNet  & \textbf{77.30} & \textbf{39.39} & \textbf{27.74} \\
    \bottomrule
    \end{tabularx}
    \vspace{-5pt}
\end{table}

\begin{table}
\caption{\textbf{Quantitative study for subject-binding attention.} Bind(1) means only augmenting the self-attention modules of the UNet down and up blocks with the smallest resolution. Conv-in refers to injecting the text embeddings through the Convolution-in layer of the reference UNet.}
\label{tab:ablationAug}
\vspace{-5pt}
\centering\scriptsize
\setlength{\tabcolsep}{5.0pt}
\begin{tabular}{lccccc}
\toprule
    Method & CLIP-I$\uparrow$ & DINO$\uparrow$ & CLIP-T$\uparrow$  & Quality$\uparrow$ & Fidelity$\uparrow$ \\
    \midrule
    w/o Binding  & 77.30 & 39.39 & 27.74  & 84 & 74\\
    Conv-in & 77.60 & 39.39 & 27.86 & 90 & 122 \\
    Bind(1) & 77.20 & 39.42 & \textbf{28.10} & \textbf{169} & 95 \\
    Bind(1,2,3) & \textbf{77.60} & \textbf{40.11} & 27.71 & 140 & \textbf{192} \\
\bottomrule
\end{tabular}
\vspace{-10pt}
\end{table}

\noindent\textbf{Subject-binding attention.}
We observe different image fidelity when binding with text embeddings on different UNet blocks, including Bind(1,2,3)~(i.e. on all blocks) and Bind(1)~(i.e. 
 on the smallest resolution). Intuitively, UNet blocks with higher resolution are likely to capture more detailed features, whereas the low-resolution blocks may primarily handle semantic information. Therefore, we ablate the selection of UNet blocks for applying attention-binding. We also try to directly inject mask map of asset image into the conv-in layer of the reference UNet.

\begin{figure}[t]
    \centering
    \includegraphics[width=1.0\linewidth]{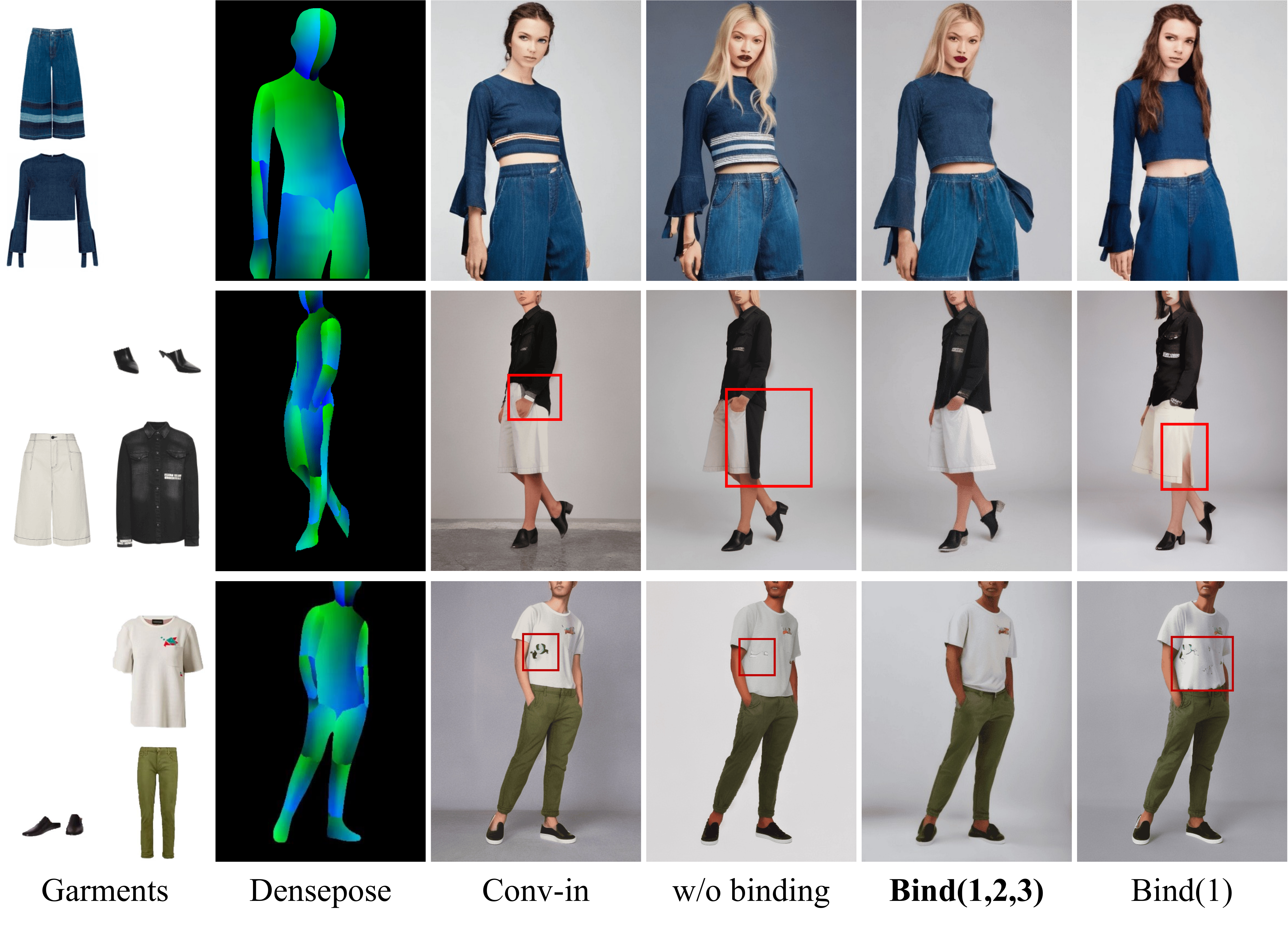}
    \vspace{-22pt}
    \caption{\textbf{Qualitative ablation study on subject-binding attention.} Bind(1) means only modifying the self-attention modules of UNet blocks with the smallest resolution. Conv-in refers to injecting the mask map through the Convolution-in layer of the reference UNet. We highlight mistakes in rows 2-3 using red boxes. }
    \label{fig: ablation_subject_binding}
    \vspace{-10pt}
\end{figure}

\cref{tab:ablationAug} illustrates the quantitative impact of subject-binding attention. Bind(1,2,3) demonstrates the best image similarities, indicating its extraordinary ability to maintain detail fidelity. Bind(1) shows worse image similarities. For the CLIP-T Score, all methods demonstrate comparable generation abilities in maintaining semantic similarity, as the quantitative values are nearly identical.
We also conduct a user study involving 23 annotators to compare the quality and fidelity of each method in \cref{tab:ablationAug}. Specifically, we randomly pick 20 compositional garment images and correspondently generate 20 images for each method.
Subsequently, we formulate two questions for each image: the first prompts the annotator to select the image with the highest quality, while the second focuses on selecting the image with the best fidelity. We aggregate the number of wins for each method in terms of quality and fidelity. Results show that injecting text representations(Bind) or mask map(Conv-in) indeed enhance the quality and fidelity of the generation results. Particularly, subject-binding attention significantly improves them, and binding text features in all UNet blocks significantly improves the detail fidelity, while restricting subject-binding attention in partial blocks sacrifices fidelity for slight improvement of quality.

\cref{fig: ablation_subject_binding} provides the qualitative ablation studies of subject-binding attention. Without subject-binding attention, the model blends the identity of different garments. 
Injecting mask maps using the convolution-in layer mitigates the subject-blending issue. However, it compromises fidelity and still cannot well discern different garments.
In contrast, subject-binding attention succeeds in distinguishing between different garments while only slightly compromising fidelity (not obvious). 
We try to mitigate this issue by restricting the subject-binding attention areas (Bind(1)), but it's more harmful to the generation fidelity.

In conclusion, given all the results, we assert that the \textit{Bind(1,2,3)} method achieves the most satisfactory balance between generation quality and fidelity in the multi-garment reference generation task.
Therefore, we leverage it as our default setting.

\begin{figure}[t]
    \centering
    \includegraphics[width=0.8\linewidth]{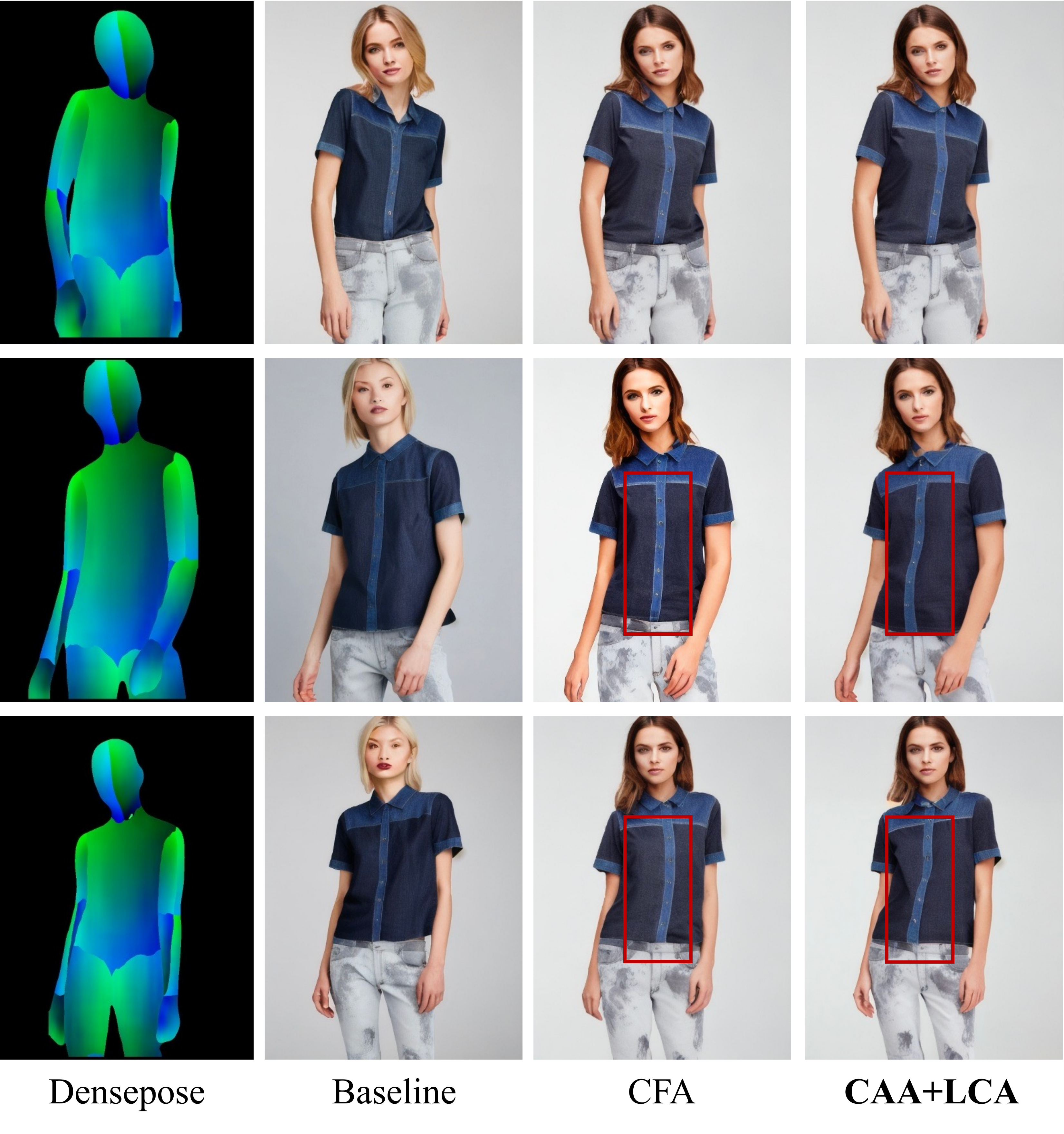}
    \vspace{-10pt}
    \caption{\textbf{Qualitative ablation study} on synthesizing independently (baseline), cross-frame attention (CFA), 
    and latent code alignment (LCA). Reference garments and text are omitted here.}
    \label{fig: ablation_consistency}
    \vspace{-8pt}
\end{figure}

\noindent\textbf{Human album generation.}
In \cref{fig: ablation_consistency}, we analyze the effectiveness of our proposed modules targeting human album generation.
In this task, we aim to generate a series of human images with the same identity.
The baseline method is synthesizing images independently with the same appearance descriptions in the text prompt. 
We observe that the baseline method generates images with totally different appearances. 
Applying cross-frame attention generates appearances with high consistency across views but compromises posture alignment (row 3) and dressing naturalness.
The model with correspondence-aware attention demonstrates better global consistency, but the appearance consistency is still not perfect. 
However, when incorporating correspondence-aware attention and latent code alignment,  our method could generate desired results with both high consistency and fidelity (row 4).

%% file: sec_arxiv/5_conclusion.tex
\vspace{-6pt}
\section{Conclusion}
We propose FashionComposer, a diffusion-based method for highly customized fashion image generation. The primary contribution is compositionality through referring to multi-modal input and multi-subject image. We propose subject-binding attention that binds visual features of various image components with their corresponding descriptions, enabling semantic distinctions between different garments while seamlessly compatible with Feature Injection Attention for fidelity preservation. However, the generation capacity of our model is still limited by the scale and bias of the training dataset in terms of race, gender, and body figure. A direct way to improve the diversity and fairness of generation is to further enhance data variety and amount.

\section{Acknowledgements}
\raggedright
This work was supported by DAMO Academy through DAMO Academy Research Intern Program.